\documentclass[runningheads]{llncs}
\usepackage{graphicx}

\usepackage{tikz}
\usepackage{comment}
\usepackage{amsmath,amssymb} 
\usepackage{color}
\usepackage{xspace}

\usepackage[accsupp]{axessibility}  

\usepackage{docmute}
\usepackage[T1]{fontenc}
\usepackage[strings]{underscore}
\usepackage{hyperref}

\usepackage{comment}

\newcommand{\Math}[1]    {{\textcolor{blue}{Math: #1}}}

\newcommand{\model}{{\sf CRISPnet}\xspace}

\newcommand\dE{\ensuremath{\Delta E^*_{ab} 76}\xspace}

\newcommand{\I}{\ensuremath{I}}

\newcommand{\rgbg}{\ensuremath{I_{RGBG}}}

\newcommand{\down}{\ensuremath{I_{D}}}
\newcommand{\fts}{\ensuremath{F}}
\newcommand{\wb}{\ensuremath{W}}

\begin{document}
\pagestyle{headings}
\mainmatter
\def\ECCVSubNumber{6126}  

\title{CRISPnet: Color Rendition ISP Net} 

\titlerunning{Color Rendition ISP Net}
%
\author{Matheus Souza \and
Wolfgang Heidrich}
\authorrunning{M. Souza and W. Heidrich}
%
\institute{King Abdullah University of Science and Technology (KAUST), \\
Thuwal, Saudi Arabia\\
\email{\{matheus.medeirosdesouza,wolfgang.heidrich\}@kaust.edu.sa}}
\maketitle

\begin{abstract}
  Image signal processors (ISPs) are historically grown legacy
  software systems for reconstructing color images from noisy raw
  sensor measurements. They are usually composited of many heuristic
  blocks for denoising, demosaicking, and color restoration. Color
  reproduction in this context is of particular importance, since the
  raw colors are often severely distorted, and each smart phone
  manufacturer has developed their own characteristic heuristics for
  improving the color rendition, for example of skin tones and other
  visually important colors. 

  In recent years there has been strong interest in replacing the
  historically grown ISP systems with deep learned pipelines. Much
  progress has been made in approximating legacy ISPs with such
  learned models. However, so far the focus of these efforts has been
  on reproducing the structural features of the images, with less
  attention paid to color rendition.

  Here we present \model, the first learned ISP model to specifically
  target color rendition accuracy relative to a complex, legacy smart
  phone ISP. We achieve this by utilizing both image metadata (like a
  legacy ISP would), as well as by learning simple global semantics
  based on image classification -- similar to what a legacy ISP does
  to determine the scene type. We also contribute a new ISP image
  dataset consisting of both high dynamic range monitor data, as well
  as real-world data, both captured with an actual cell phone ISP
  pipeline under a variety of lighting conditions, exposure times, and
  gain settings.

\keywords{image signal processor; image restoration; color rendition.}
\end{abstract}

\begin{figure}[h]
\centering
\includegraphics[width=1.0\linewidth]{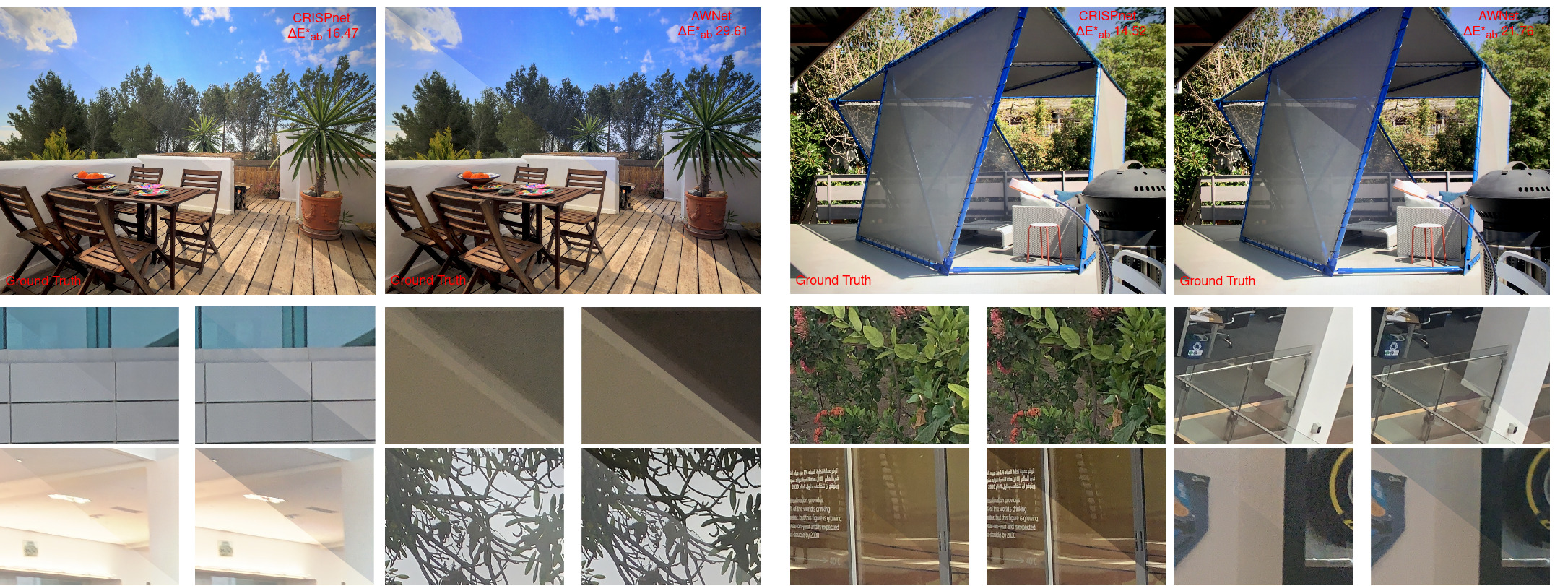}
\caption{Examples of faithful color rendition produced by \model
  compared to AWNet~\cite{awnet} for two full images and several image
  patches. The lower triangle in each image is the ground truth, the
  upper triangle is the respective learned model. Within each pair,
  the left result is \model, the right AWNet.
  Note also the
  quantitative improvement in \dE (lower is better).}
\label{teaser}
\end{figure}

\section{Introduction}

The past decade has seen tremendous progress in miniaturizing camera
modules to fit high quality imaging systems into flat mobile
devices. As this miniaturization effort is approaching physical
limits, such systems increasingly rely on computational methods to
maintain high image quality. Image Signal Processors (ISPs) are
responsible for tasks such as interpolation, demosaicking, denoising,
enhancement on the edges, white balancing and color restoration,
exposure correction, gamma encoding, compression, and so forth.  The
increasing miniaturization places a higher burden on these tasks, and
in particular the reproduction of accurate colors. For example, the
pixel pitch in modern smart phone cameras is now as small as 0.7
$\mu$m -- barely larger than one wavelength of red
light\footnote{e.g. Samsung ISOCELL sensors}, leading to significant
crosstalk between the pixels in a Bayer pattern
image~\cite{anzagira2015color}. This distortion is a spectral effect
that can not be accurately modeled as a simple deconvolution or color
matrix in RGB space.

Existing ISPs combat such image artifacts with complex, historically grown
pipelines made up of heuristic blocks that are prone to error
accumulation through the pipeline. At the same time, these systems are
based on substantial expertise in color theory and human perception,
and lead to a brand-specific ``look'' that accounts for a lot of the
appeal of particular smart phone models.

In recent years there have been a number of proposals for replacing
the historically grown and heuristic ISP pipelines with simpler, more
principled approaches. Initially, these efforts relied on
optimization-based approaches~\cite{heide}, but with increasingly more
powerful compute resources~\cite{smart} the attention has now shifted
to deep learning and convolution neural networks
(CNNs)~\cite{ratnasingam2019deep,Schwartz2019}. More recent efforts
include PyNET~\cite{ignatov2020replacing} and similar
methods~\cite{csanet,pynetca,awnet} that train ISP networks to match
DSLR data.

We believe this approach is fundamentally limited in that {\bf DSLRs
  can not be considered the gold standard in color processing}: while
DSLRs have superior optics and noise characteristics, they have very
minimalistic ISPs, and effectively offload the problem of color
rendition to the end user -- typically a professional photographer who
will manually post-process the images in photoshop or similar tools.

In this work we therefore propose a learned ISP framework that
leverages in particular the color science expertise that has gone into
the development of legacy smart phone ISPs. Color Rendition ISP Net
(\model) is trained on pairs of raw and ISP-processed smart phone
images. In addition, it also learns to leverage the same white balance
metadata that also informs the legacy ISP pipelines. Finally, like
many legacy ISPs, \model performs a rough semantic image analysis, for
example to distinguish between landscape and portrait shots. We show
that each of these architectural improvements significantly increases
reconstruction accuracy, especially with respect to color rendition.
In summary, we make the following contributions:



\begin{itemize}
\item a novel deep architecture for ISP processing specifically
  tailored at matching the color processing of legacy ISPs.
\item a way to inject white balance metadata into the network so that
  it can learn to be robust under different illuminations and scenes.
\item an attention-based image semantics module that helps the
  ISP net to make different decisions for different scene types.
\item an extensive new dataset for ISP networks consisting of pairs of
  raw and ISP processed smart phone camera images over a wide range of
  exposure and gain settings. The dataset consists of two parts -- a
  large monitor dataset captured from a color calibrated HDR monitor
  (used for large scene diversity), and a smaller real-world dataset
  used for fine tuning.
\end{itemize}

Code and data will be made available.

\section{Related Work}

The idea of treating all stages involved in an ISP pipeline as a
single integrated problem was proposed in \cite{heide}. Their
framework takes the traditional modular structure applying different
heuristics in each stage and substituted it with a single inverse
problem. The proposed approach uses proximal operators and the
primal-dual method for optimization, performing all of that together
with various natural image priors (BM3D \cite{BM3D}, TV, and
cross-channel \cite{cross}). However, these natural image priors are
relatively weak compared to modern deep learned models, leading to a
degradation in image quality.

Later the optimization strategies were substitute by deep learning
methods, building on excellent results reported for in image
processing tasks like denoising \cite{d1,d2,d3,d4}, deblurring
\cite{db1,db2,db3}, super-resolution \cite{sr1,sr2,sr3,swinir},
etc. These systems focused on solving denoising and demosaicking
jointly, learning mappings from raw to sRGB and from sRGB to raw
\cite{pseudoisp,cycleisp}. Although these works can diminish noise,
the color fidelity is sometimes left behind. 

Other works on deep learned ISPs were focusing on image enhancements.
Schwartz et al.~\cite{Schwartz2019} proposed a DeepISP mapping from a
low-light mosaicked image to final sRGB, while Igantov et
al.~\cite{ignatov2020replacing} proposed PyNET, targeting translation
of mobile raw images to match the quality DSLR sRGB. After PyNET, more
architecture ideas emerged~\cite{contest2020,contest2021}, the most
competitive ones relying on encoding-decoding structure with backbone
resembling network like ResNET, UNet, Pix2Pix,
etc.~\cite{resnet,unet,pix2pix} often combined with attention
mechanisms~\cite{csanet,pynetca,awnet}, these models further increase
the accuracy, exploiting global and local features, applying spatial
and channel attention similar to~\cite{cbam} proposed.

However, all these works have adopted the notion of training a mapping
between mobile raw data and DSLR color images. While DSLRs have
superior optics and sensors, they mostly rely on the user to
post-process the image to improve the color rendition, while the ISPs
of smart phones and point-and-shoot cameras are optimized to produce
good looking final color images without manual post-processing. As a
result many consumer camera tests now consider mobile camera images to
have better color rendition than DSLRs without
post-processing.\footnote{e.g. https://www.photographyacademy.com/why-phones-take-better-pictures-than-your-dslr}

In this work we therefore aim to leverage the substantial expertise in
color theory and human perception that has flown into legacy ISP
pipelines, and train a network to reproduce the color processing of a
legacy smart phone ISP. Since such ISP pipelines make many non-linear
and scene-dependent decisions, we show that this mapping can not be
learned effectively with previous ISP network architectures. We
therefore propose two architectural improvements: first, we utilize
image meta data captured with the raw data.  Image meta data was
previously used in tone mapping applications~\cite{meta}, but it also
is used extensively in traditional ISP pipelines. In this work, we
specifically usilize white balance information to learn an ISP network
with improved color rendition. We note that the use of white balance
information differs form the previously proposed use of the color
matrix metadata~\cite{cameranet}, since the color matrix is a
scene-independent characteristic of the camera hardware, while the
white balance data is scenes-specific.  We also propose a global
feature learning approach to feed scene composition and semantics
information into the ISP process. For this purpose, we adopt the new
XCiT transformer architecture~\cite{xcit}, which we found to work
better for this purpose than alternatives like Swin
transformers~\cite{swin}.

As mentioned above, the most commonly used dataset for ISP networks
uses pairings of mobile and DSLR images~\cite{ignatov2020replacing},
and is therefore not of use for our system. Other existing datasets
(e.g.~\cite{Schwartz2019}) are too small to really benefit our
method. We therefore also propose a new dataset with a mix between HDR
monitor captures and real world data. The HDR monitor data provides
large scene diversity with excellent dynamic range and color
calibration but is limited by the resolution of the monitor, which is
lower than that of the camera. The real world data is used for fine
tuning the recovery of pixel-level details, but cannot match the scene
diversity of the monitor data.

\section{Method}
\begin{figure}[h]
\centering
\includegraphics[width=1.0\linewidth]{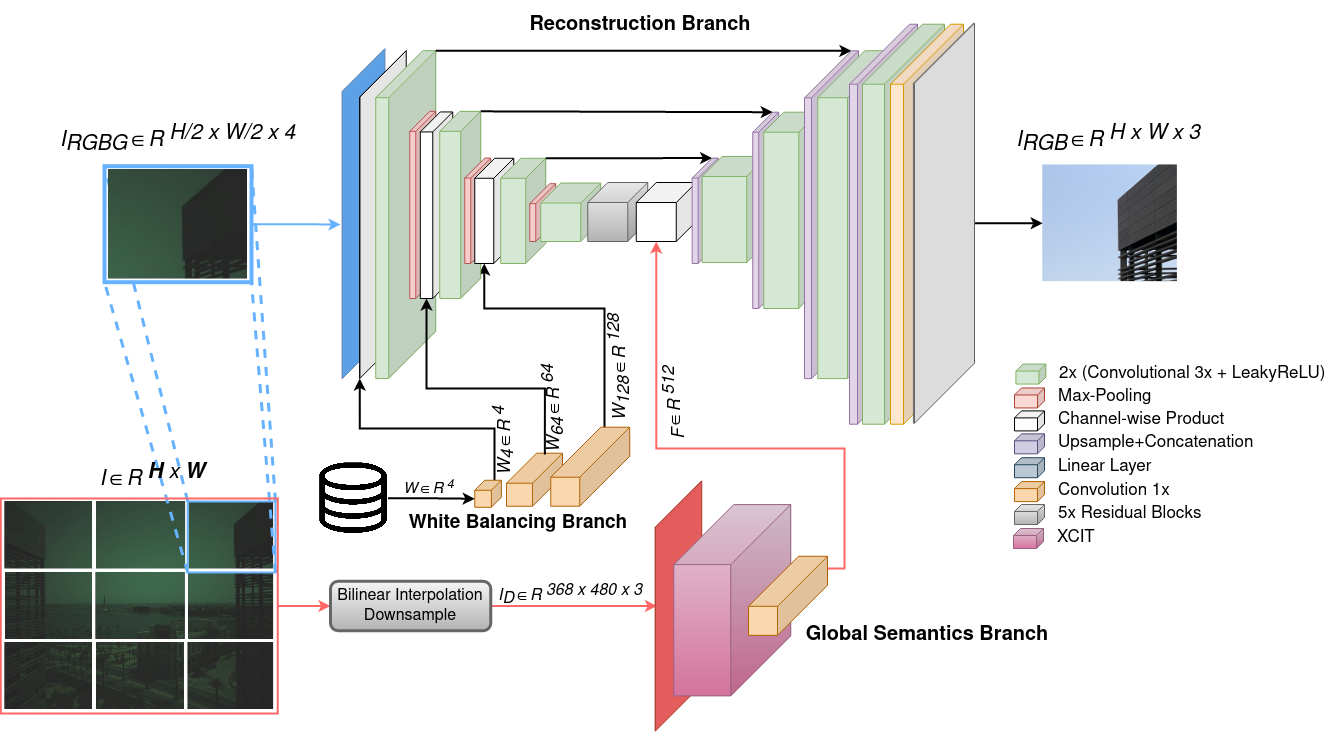}
\caption{An illustration of the \model. It receives as input the image
  divided into patches, the full image downsampled to 368 x 480
  pixels, and the white balancing metadata. Here, \textbf{H} and
  \textbf{W} represent the full image dimension, while H and W is the
  dimension of the image patches.  \wb represents the white balance
  value for each image channel. The network is divided into three
  branches. The reconstruction branch is responsible for overall image
  reconstruction while aggregating information from the other branches
  efficiently. The white balance branch projects the white balance
  information to match the layer dimension inside the reconstruction
  and the global semantics branch learns global scene semantics from a
  downsampled version of the full image. a) and b) correspond to two
  different strategies explored in this work.}
\label{arch}
\end{figure}

The ISP problem can be interpreted as an image translation task. Given
a raw image, we want to learn a mapping to RGB that maintains the
characteristics of the original device. Therefore, during this
process, the method must learn low-level and high-level properties
such as denoising, deblurring, sharpening, white balance, etc. The
recent ideas provide end-to-end solutions mainly focusing on denoising
and demosaicking \cite{cycleisp,pseudoisp,Schwartz2019}. However, in
modern ISP pipelines, the white balance and the global scene
classification heavily influence the final result and will be explored
in the following.

\model follows the architecture shown in \figurename{\ref{full}}.
The Bayer raw image $\I \in\mathbb{R}^{H \times W}$ goes through two
branches. The reconstruction branch restructures the Bayer mosaic $\I$
into a 4 channel image $\rgbg \in\mathbb{R}^{\frac{H}{2}\times
  \frac{W}{2} \times 4}$. Tiles of this image are processed with a
modified UNet, which is also responsible for aggregating information
from the other branches. The output of the reconstruction branch are
ISP processed tiles that are reassembled into a translated RGB image
$I_{RGB} \in\mathbb{R}^{H \times W \times 3}$.

The second branch operates on a downsampled version of the full image
$\down \in\mathbb{R}^{480 \times 368}$ with fixed resolution. This
branch is responsible for learning global semantics when the training
images are divided into patches. The output of this branch $\fts
\in\mathbb{R}^{512}$ matches the dimensions of the bottleneck in the
reconstruction branch and the two branches are combined with a
channel-wise product.

Finally, we also inject image meta data into the reconstruction branch
at three separate stages, again through a channel-wise product. The
meta data consists of white balance information -- three channel weights that
are processed through a third network branch. We also experimented
with additional meta data such as exposure time and gain, but did not
find them useful in improving the results. On the other hand, a
learned injection of the white balance data does show significant
improvements compared to both not using meta data at all and also
compared to simple pre-multiplication of the white balance weights on
the input data (see Section~\ref{sec:exp}).

\subsection{Reconstruction Branch}

The backbone of \model has a UNet structure \figurename{\ref{full}},
which has proved to be effective for this task by many of the works
proposed in \cite{contest2020,contest2021}. Its fully
convolutional structure enables efficient inference for high
resolution images. Like other learned ISPs, we perform patch-based
image processing in order to control the computational expense for
training the network.  Our implementation has three main differences
when compared with ``vanilla'' UNet proposed by \cite{unet}. Here we
want to aggregate additional information in order to reconstruct
faithful colors.

In legacy ISP pipelines white balance information is taken into
account in several stages. To reproduce this behavior, we aggregate
this information in the early layers using the white balance branch
output as scales for the first three downsample levels in the
network. Different ideas were proposed to better perform feature
matching, including concatenation, dot-product, as well as more
complex solutions like \cite{film}. However, as detailed below, since
the white balance branch is based on a small number of inputs, a
simple projection works best for this purpose.

After aggregating the ambient light information exploiting white
balance data, we add residual blocks in the bottleneck. Mixing ideas
from ResNET and UNet is not new; \cite{2020,deepunet} explored the
power of both architecture together. In this work, we also exploit
this, using the residual blocks only in the network bottleneck, as
\figurename{\ref{full}} illustrates. For our application we observed
that residual blocks in the encoder part lead to overfitting.

Finally we inject global features from the scene learned by the
implicit global semantics branch. We take advantage of the
compressed representation after the residual bottleneck to match the
global features $\fts \in \mathbb{R}^{512}$. We experimented with a
number of injection strategies, but finally adopted a channel-wise
product, which scales the information inside the bottleneck channels,
increasing or diminishing the impact of feature channels in the final
reconstruction.

\subsection{White Balance Awareness}
The raw file contains unprocessed or minimally processed data as well
as image metadata that describes image characteristics and parameters
chosen by the device during the shot, based on light conditions or
user preferences. Traditional raw to RGB solutions take advantage of
this data but most current Deep Learning models ignore it.

The standardized metadata in a DNG file includes parameters such as a
color matrix, white balance weights, as well as exposure and ISO
settings. Of these, the color matrix describes the camera hardware but
not the specific image; as such it is easy to implicitly pick up for a
neural network just based on image pairs without explicitly using the
metadata information. Similarly, we experimentally found that ISO and
exposure settings do not improve the ISP reconstruction task. The
situation however is very different for the white balance data -- here
we notice a substantive improvement by injecting this type of metadata
into the learning process.

The white balance meta data comes in the form of one scalar multiplier
per color channel, where the green channel is usually normalized to a
value of 1. We expand this information to $\wb \in \mathbb{R}^{4}$ in
RGBG format. Learned convolutional layers then upsample this
information to the same size as the corresponding layers in the
reconstruction branch so that the white balance information can be
injected into the reconstruction branch at three distinct locations
with a channel-wise dot product.

Note that the white balance branch does not contain any activation
functions -- since the input is only three scalars, and no scene
information is available, we did not find it helpful to include
non-linearities in this branch. Instead, the information is simply
injected into the reconstruction branch, which can then learn how to
best merge the information from the image and the meta data.

In Section~\ref{sec:exp} we show that this {\em learned} utilization
of metadata is superior to the naive approach of just pre-multiplying
the image with the three scalars. We also show in the ablation studies
that three injection points are a sweet spot to balance image quality
and network complexity.


\subsection{Global Semantics Branch}
Unlike the ISPs in DSLRs, the legacy ISPs in mobile phones and
point-and-shoot cameras perform highly scene dependent image
processing to produce good looking images without manual
intervention. Different scenes (e.g. indoor vs. outdoor, portraits
vs. landscapes) and conditions (e.g. sunny vs. overcast, snowy
vs. green) may require different color adjustments to produce
appealing final images. Reproducing this behavior is extremely hard
because ISPs are closed software, and is not straightforward to infer
how all the high level and low level image processing blocks work
inside.

Deep learning models have the power to learn global semantics, and
attention mechanisms have proven to work particularly well for this
purpose. Previous works \cite{awnet,csanet,pynetca} apply attention
mechanisms to capture the relation between objects, however since they
work on patches, they have no access to the global scene semantics.
To address this issue, \model has a specialized branch to deal with
global semantics, exploiting the efficient transformer architecture
XCiT \cite{xcit} to extract and combine global features We also
discuss alternative ideas that we experimented with.

Different from the reconstruction branch, this branch takes as input the full raw image $\I \in\mathbb{R}^{H
  \times W \times 4}$, downsampled to $\down \in\mathbb{R}^{368 \times
  480 \times 3}$. While the main reconstruction branch is
convolutional and can operate either on patches or full frame images,
this branch therefore always operates on images of the same size, and
has full access to the global semantics of the scene independent of
patch cropping.

This empowers \model to exploit global scene semantics in the ISP
reconstruction task. Specifically, we propose to exploit attention
mechanisms is a different way from previous approaches with spatial and
channel attention \cite{csanet,pynetca}. Here, we apply a
transformer-based approach.  Transformers are known for their large
receptive field and hence the ability to learn global
representations. The recent Cross-Covariance Image Transformers (XCiT)
\cite{xcit} has these benefits while remaining efficient; we therefore
chose them as the core of our global feature branch.

The XCiT transformer, instead of having all tokens attending to all
tokens, applies a ``transposed'' version of the ViT idea \cite{vit},
where the features channels attend to other channels, making the
complexity linear in the image resolution.
In XCiT, the attention is computed using the cross-covariance between
queries and key projections of tokens. This is motivated by the
relationship between the Gram matrix and the covariance matrix, in
which the eigenvalues of one can be obtained by decomposition of the
other.
This cross-covariance self-attention is followed by a local patch
interaction and a feed-forward network.
In \model, the ``tiny'' version of XCiT was  simplified even more,
reducing its depth to 4 -- legacy ISP classification strategies are
simple, and we did not want to over parameterize this branch.

The downsampled full image passes through these blocks, and the final
representation is then aggregated using the final hidden state of the
CLS token~\cite{cls}, which is commonly used for classification tasks.
This sequence is projected using a 1x1 convolutional layer to match
the bottleneck dimension of the reconstruction branch
$\fts\in\mathbb{R}^{512}$. Finally, the features are combined
using a channel-wise product.

This pipeline produces consistent results between patch-based and
full-frame inference.  As Section~\ref{sec:exp} shows empirically,
learning full image global representations substantially improves the
reconstruction accuracy.

In earlier versions of \model, we also experimented with alternative
ways to extract global semantics. For example, we tried an
architecture resembling a simple classification network, where the downsampled raw image passes
through strided convolution followed by batch normalization and ReLU,
and then a max pooling layer. This process was repeated twice and
lastly, a fully connected layer was applied to encode the information
to $\fts \in\mathbb{R}^{512}$. We matched this compressed
representation with the reconstruction branch bottleneck using a
channel-wise product. Notice that we did not downscale the feature to
the dimension of the label because they are not known. Instead, we
implicitly learned the global features for the current image through
the same loss function as the reconstruction branch. While this
approach worked well overall, we observed more issues with noisy low
light images, and instead adopted the XCiT model described above.

\section{Dataset and Training}

\subsection{Dataset}

To train \model, we require a large dataset of pairs of raw and
ISP-processed mobile phone data. Existing datasets are not suitable
for this task since they are either too small for our
purposes~\cite{Schwartz2019}, or use a DSLR as a reference
camera~\cite{ignatov2020replacing}.

We therefore captured our own dataset. Since global semantic scene
information is crucial to our approach, we require a large diversity
of different scene types. We therefore resort to a two-part dataset: a
large database of monitor-captured images that that cover a wide
range of different environments, including indoor and outdoor, and
different types of landscapes. In addition, we also capture a smaller
real-world dataset that is used for fine-tuning to overcome any pixel
artifacts that may occur due to the monitor data.

All data was captured using an iPhone XR and the ProCam software for
IOS. Every shot using this app generates two images of size
$4032\times3024$: raw (DNG format) and RGB. The raw is a single
channel grid (Bayer Pattern) with 16 bit values, which represents the
measured light intensity. The RGB images are generated by passing
the same raw images through the iPhone XR pipeline.

\paragraph*{Monitor data.}
The monitor captures were performed in a dark room using the CG3145 4K
HDR monitor. This monitor has a typical contrast ratio of
1,000,000:1, which it achieves with a dual modulation
principle~\cite{seetzen2004high}.  However, unlike most most dual
modulation HDR monitors that use an LCD illuminated by a low-frequency
LED backlight~\cite{seetzen2004high}, the Eizo CG3145 actually uses
two LCD layers stacked on top of each other on the same glass
substrate. This allows the monitor to achieve not only high global
contrast, but also excellent local contrast of high frequency
features. As such, it is capable of producing high quality images over
a large range of intensities, as well as individual images that exceed
the dynamic range of the mobile phone camera to simulate challenging
illumination conditions. The monitor was color calibrated using off
the shelf software for accurate representation of the source material.

Using a tripod-mounted phone setup and a capture script, we acquired a
total of 2000 raw/ISP image pairs.  The source material was taken from
\cite{flk} as well as from a new HDR portrait
dataset.\footnote{Details will be provided in the final version to
  preserve double blindness.}

\paragraph*{Real world data.}
One downside of the monitor setup is that the screen resolution of
$4,096 \times 2,160$ is lower than the resolution of the mobile phone
camera, which introduces pixel level artifacts. To combat these, we
also capture a real world dataset for fine-tuning the training. This
dataset has accurate pixel-level details, but much lower scene
diversity than the monitor data, since it is not feasible to travel
to remote places to capture images in different sites. All in all, we
captured 200 real world image pairs.

\subsection{Training}

From the monitor dataset we select $750$ images ($600$ for training,
$75$ for testing, and $75$ for validation) and from the real-world
dataset $198$ images ($160$ for training, $19$ for testing, and $19$
for validation).  The idea here is to first train using only monitor
data and then fine-tune with the real-world. A small crop was
performed on the monitor data to avoid capturing the bezel of the
monitor itself, reducing the resolution from $4032\times3024$ to
$3840\times2944$. Next, each picture was divided into $64$ patches of
$368\times480$ to train the reconstruction part. While the
reconstruction receives a patch, the global feature acquisition branch
is fed with the downsampled as discussed above.

\subsection{Loss Function}

In our experiments we adopted the MSE loss function. Usually in this
field, ``perceptual'' and SSIM losses are used together with MSE or
MAE. However, we did not observe sufficient improvements from these
losses to justify the increased training overhead. We also note that
these alternative losses focus on structural image features and tend
to neglect color reproduction.  In the supplementary material we
provide a more detailed analysis.

\section{Experiments}
\label{sec:exp}

\begin{figure}[h]
\centering
\includegraphics[width=0.7\linewidth]{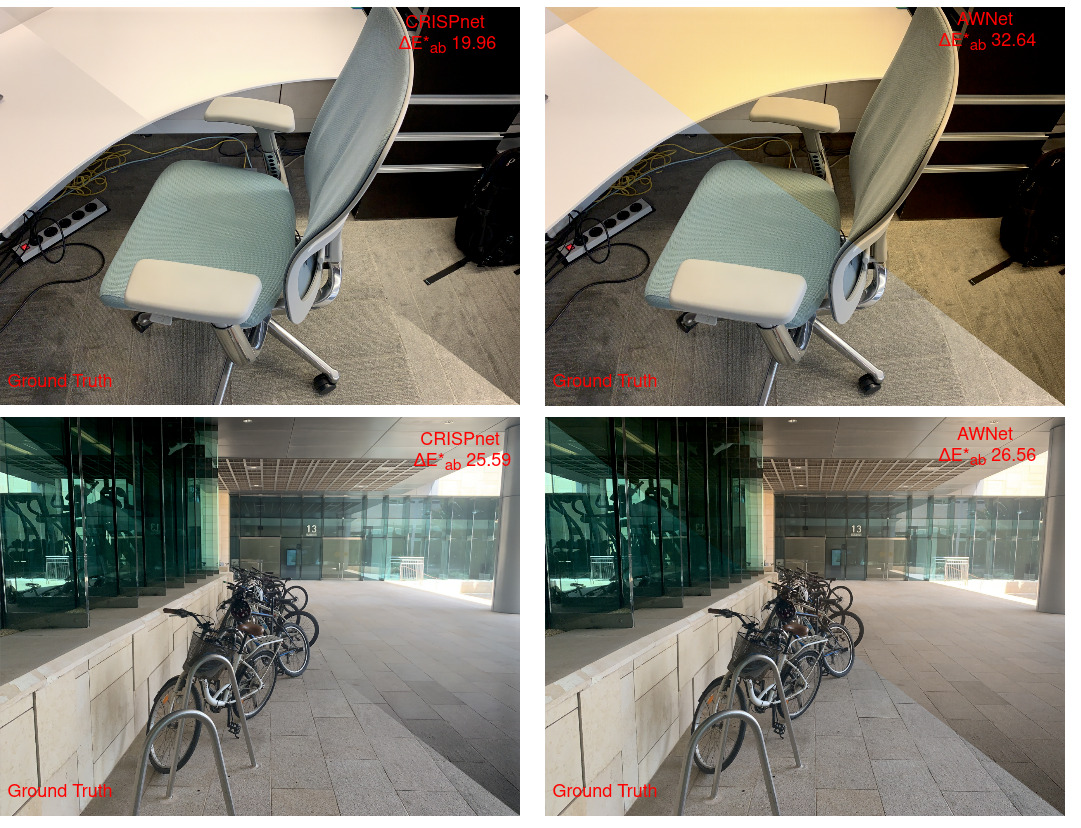}
\caption{Full frame image reconstruction after fine-tuning \model to deal with real-world data, compared against AWNet \cite{awnet}. \model is able to reproduce the light condition in the scene and threfore producing accurate colors.}
\label{full}
\end{figure}



For comparison with the state-of-the-art, we re-train several SOTA
methods on the same dataset described above.
The chosen methods we compare with are AWNet \cite{awnet}, PyNET CA \cite{pynetca} and CSANet \cite{csanet}. All these strategies are improvements over PyNET \cite{ignatov2020replacing} and the best reported results in Mobile ISP contests \cite{contest2020,contest2021}. All three methods were trained using our dataset following the technical recommendations, parameters, and loss functions described by them. AWNet and PyNET CA provided code and CSANet was implemented by ourselves. The strategy of training with monitor data and then using the real-world to fine-tune was performed for them as well. \model is also compared against the UNet with residual bottleneck, which is our network backbone without the global feature acquisition and white balance branches. 

To assess the color rendition accuracy we use the \dE metric, which is the RMS error in the non-linear, {\em perceptually uniform} CIELAB space. Perceptual uniformity means that the distance in the CIELAB space can be directly mapped to {\em just noticeable differences} (JND) between two similar colors. Specifically, one JND corresponds to $\dE\approx2.3$, so that two colors that differ by this value can just barely be distinguished by a standard human observer. We also adopted the standard PSNR and SSIM metrics.

\begin{table}[h]
\vspace{-0.2cm}
\centering
\begin{tabular}{l|r|r|r}
\hline
\multicolumn{1}{|l|}{Method}                 & PSNR $\uparrow$ & \dE  $\downarrow$ & \multicolumn{1}{r|}{SSIM $\uparrow$} \\ \hline
CSANET \cite{csanet}                                      &     24.51   &   16.07  &  0.8631                         \\
PYNET CA  \cite{pynetca}                                      &  25.00 & 15.50    &  0.8561                           \\
AWNET \cite{awnet}                                      &   25.70   &   14.41  &    0.8947                       \\
UNet + Residual Bottleneck                                &  26.33    &  13.76   &    0.9130                       \\
\model      &  27.54    &  11.75   &  0.8911                         \\ \hline
\end{tabular}
\caption{Reconstruction results on Real-World dataset first fine-tuning using Monitor data.}
\label{tab:fine}
\end{table}

\vspace{-0.8cm}

Table \ref{tab:fine} shows quantitively our final result evaluated
with real-world data after fine-tuning. \model provides a substantial
increase in terms of PSNR and \dE over our comparisons. We can observe
how the other methods suffer to translate faithful colors and how
relying only on PSNR and SSIM does not reflect the color rendition. In
particular the SSIM is too focused on structural similarity and
overlooks color discrepancies that are very noticeable in
\figurename{\ref{full}}.
When only monitor data training is taken into account the metrics are higher overall (Table~\ref{tab:monitor}). For this case, we have a larger dataset, with captures done always in the same camera positioning settings, and the monitor resolution is lower than the camera, these facts make the reconstruction task easier. \model proves to better exploit these characteristics and outputs better results in all evaluated metrics. \figurename{\ref{teaser}} shows qualitatively that our model is almost indistinguishable from the Ground Truth, while AWNet cannot translate colors precisely.

\begin{table}[h]
\centering
\begin{tabular}{l|r|r|r}
\hline
\multicolumn{1}{|l|}{Method}                 & PSNR $\uparrow$ & \dE  $\downarrow$ & \multicolumn{1}{l|}{SSIM $\uparrow$} \\ \hline
CSANET  \cite{csanet}                                      &    27.05 &  11.43   &   0.8986                        \\
PYNET CA \cite{pynetca}                                     &   27.41   & 11.06    &  0.8796                          \\
AWNET \cite{awnet}                                         &   27.66   & 10.84    &  0.9093                         \\
UNet + Residual Bottleneck                               &   28.34   &  10.25   &  0.9243                         \\
\model          &  32.04    &  6.36   & 0.9344                          \\ \hline
\end{tabular}
\caption{Reconstruction results on Monitor dataset. This is considerer in our pipeline a pre-training step to finally fine-tune for Real-World data.}
\label{tab:monitor}
\end{table}

\subsection{Ablation Studies}

\begin{figure}[h]
\centering
\includegraphics[width=1.0\linewidth]{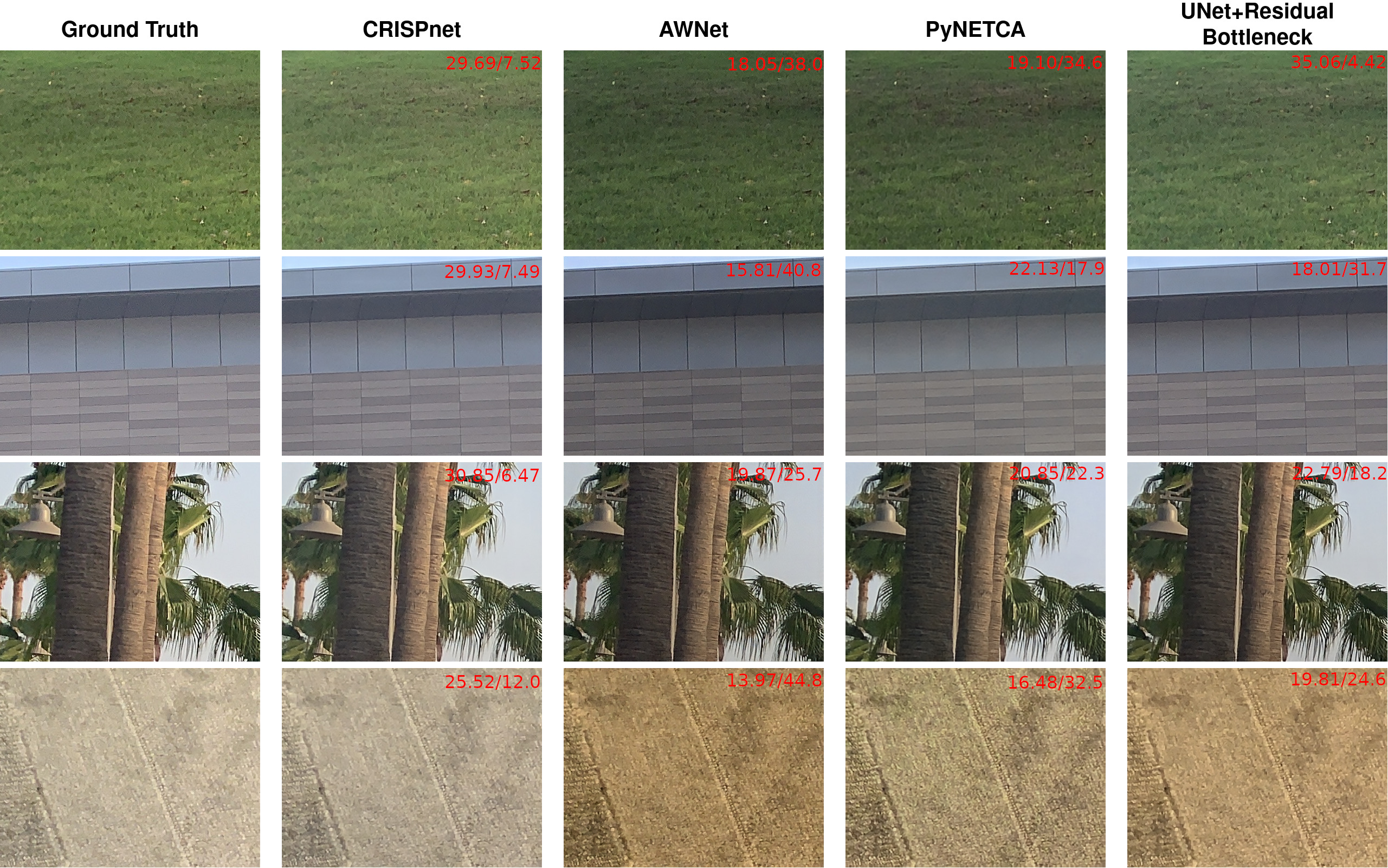}
\caption{\model compared against AWNet, PyNET CA and our backbone (UNet+Residual Bottleneck). PSNR/\dE.}
\label{all}
\end{figure}

Tables \ref{tab:fine} and~\ref{tab:monitor} show how each proposed
improvement enhances the reconstruction process. Initially, we have a
well-known UNet \cite{unet} with residual connection in the
bottleneck. Next, we introduce the white balance (WB) information,
Table \ref{tab:wb} ablates over not using metadata at all, naively
using white balance information as preprocessing step, and finally
using the proposed branch with the UNet with residual bottleneck. We
can observe quantitatively its importance and \figurename{\ref{all}}
clarify this even more. Through Table \ref{tab:wb} we empirically
justify our design choice of injecting white balancing information at
three points. We also show that preprocessing the images with metadata
for AWNet \cite{awnet} is not effective as our proposed branch --
Table~\ref{tab:wb} makes this clear.

\begin{table}[h]
\centering
\begin{tabular}{l|r|r|r}
\hline
\multicolumn{1}{|l|}{Method}                 & PSNR $\uparrow$ & \dE  $\downarrow$ & \multicolumn{1}{r|}{SSIM $\uparrow$} \\ \hline
AWNet \cite{awnet}                                 &  27.66   & 10.84 &  0.9093                         \\
AWNet \cite{awnet} Preprocess WB                   &  28.51   & 9.88  &  0.9113                        \\\hline
No WB					                           &  28.34   & 10.25 &  0.9243                         \\
Preprocess WB 									   &  28.98   & 9.57  &  0.9236                        \\
Branch level 1               					   &  29.84   & 8.42  &  0.9261                       \\
Branch level 1, 2 and 3      					   &  29.86   & 8.40  &  0.9281                       \\
Branch level 1, 2, 3 and 4           			   &  29.75   & 8.48  &  0.9264                       \\ 
Branch level 1, 2, 3, 4 and 5          			   &  29.69   & 8.56  &  0.9256                       \\ \hline
\end{tabular}
\caption{Comparison of different ways to explore the White Balance (WB) information. AWNet is slightly improved by preprocessing the input with white balance metadata. However, it still behind our results even before the global semantics branch been added.}
\label{tab:wb}
\end{table}

We also ablate about the importance of training first with monitor
data and then fine-tune with real-world. The  additional branches of \model
make it powerful enough to learn color conditions that can be heavily
refined during fine-tuning. If we train \model only with monitor data
and reconstruct real-world scene our PSNR, for instance, stays around
23.9 dB, after fine-tuning it increases beyond 27.5 dB. Table~\ref{tab:real}
shows what happens when we train straight using real-world data for
both \model and the UNet with residual bottleneck. Comparing against
Table \ref{tab:fine} we can notice that for all metrics the results are
degraded. Therefore, our proposed dataset attached with the fine-tune
strategy enables better output without the requirement of taking
thousands of real-world image pairs.

\begin{table}[]
\centering

\begin{tabular}{l|r|r|r}
\hline
\multicolumn{1}{|l|}{Method}                 & PSNR $\uparrow$ & \dE  $\downarrow$ & \multicolumn{1}{r|}{SSIM $\uparrow$} \\ \hline
UNet + Residual   &  26.03   &  14.14   &  0.9038                         \\
\model            &  26.73    &  12.51   & 0.8746                        \\ \hline
\model Fine-tuned             & 27.54    &  11.75   &  0.8911                         \\ \hline
\end{tabular}
\caption{Result on the Real-World data without and with fine-tunning process.}
\label{tab:real}
\end{table}

Finally, we observe the impact of global semantics branch and the
different ways to do it. \model extracts the global semantics
information through an XCiT transformer model \cite{xcit}, which is a
more powerful attention mechanism than traditional spatial or channel
attention. Table~\ref{tab:gfa} shows quantitatively that exploiting
global semantics indeed improves the reconstruction quality and a
light transformer-based attention mechanism beats a simple CNN
approach for this purpose. More ablations about different attention
mechanisms and how \model performs with deeper XCiT models are
available in the supplementary material.

\begin{table}[]
\centering
\begin{tabular}{l|rrr|rrr}
\cline{2-7}
                                       & \multicolumn{3}{l|}{Monitor}                                                         & \multicolumn{3}{l|}{Real-World}                                                       \\ \hline
\multicolumn{1}{|l|}{Global Semantics} & \multicolumn{1}{l|}{PSNR$\uparrow$}    & \multicolumn{1}{l|}{\dE  $\downarrow$}  & \multicolumn{1}{l|}{SSIM$\uparrow$} & \multicolumn{1}{l|}{PSNR$\uparrow$}    & \multicolumn{1}{l|}{\dE  $\downarrow$}   & \multicolumn{1}{l|}{SSIM$\uparrow$} \\ \hline
Without                                & \multicolumn{1}{r|}{29.86} & \multicolumn{1}{r|}{8.40} & 0.9281                    & \multicolumn{1}{r|}{26.80} & \multicolumn{1}{r|}{13.01} & 0.9133                    \\
Classification CNN                     & \multicolumn{1}{r|}{31.89} & \multicolumn{1}{r|}{6.55} & 0.9311                    & \multicolumn{1}{r|}{26.90} & \multicolumn{1}{r|}{12.62} & 0.8874                    \\
XCiT                                   & \multicolumn{1}{r|}{32.04} & \multicolumn{1}{r|}{6.36} & 0.9344                    & \multicolumn{1}{r|}{27.53} & \multicolumn{1}{r|}{11.75} & 0.8911                    \\ \hline
\end{tabular}
\caption{Comparison between different global semantics strategies and its contribution.}
\label{tab:gfa}
\end{table}

\section{Discussion, Limitations, and Future Work}

In this work we introduced \model, the first ISP network designed to
improve color rendition by learning from the expertise encoded in
legacy ISP pipelines. We achieve this by combining a convolutional
encoder/decoder architecture for the main reconstruction task with
both white balancing metadata and a transformer-based global feature
branch. We demonstrate substantial improvements in both traditional
image metrics (PSNR, SSIM), as well as color accuracy (\dE).

\begin{figure}
  \vspace{-2mm}
  \centering
  \includegraphics[width=0.7\linewidth]{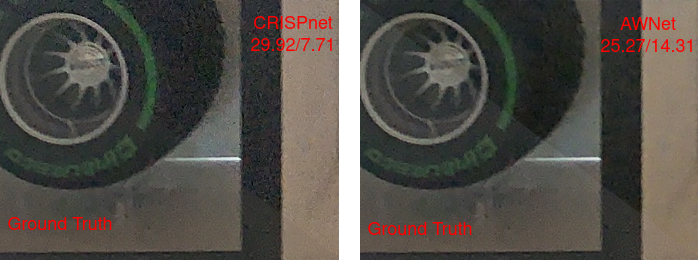}
  \vspace{-2mm}
  \caption{For very low light images, our method can have more issues
    with noise than AWNet. However, the significant improvement in
    color rendition still results in an overall increased PSNR for
    \model. PSNR/\dE. \label{fig:noise}}
  \vspace{-2mm}
\end{figure}

However, the proposed approach is not without shortcomings. As can be
seen from both the quantitative and the qualitative results, the color
rendition is much improved, but often still not below the noticeable
threshold. This indicates that legacy ISPs are still more complex than
the existing network architectures are capable of reproducing.
Furthermore, while our architecture improves color rendition, it can
have sometimes have issues with noise especially in very low light.
As shown in Figure~\ref{fig:noise}, the color improvements usually
still outweigh the poorer noise performance for an overall increased
PSNR.

Despite the low-light issue we believe that our approach is preferable
in many situations -- global color distortions are often significantly
more noticeable than fine scale noise in typical use cases for the
images (e.g. when the images are posted on social media in lower
resolution anyways).

In the future it would be interesting to explore how to improve the
noise performance, for example by having both a DSLR reference image
to improve the structural details of the image and a mobile phone ISP
image for the color rendition. Ultimately it would be best if we did
not require a reference ISP algorithm at all and could instead learn
to reproduce the manual color adjustments made by skilled professional
photographers.  We believe the architectural changes we propose in
this paper can also be useful for these types of systems.

%
\bibliographystyle{splncs04}
\bibliography{egbib}

\end{document}